\documentclass{article}
\usepackage{spconf,amsmath,graphicx}
\usepackage{breqn}
\usepackage{bbm}
\usepackage[font=small,labelfont=bf,skip=4pt]{caption}
\usepackage{romannum}
\usepackage{amsfonts}
\usepackage{ctable}

\usepackage{booktabs}
\newcommand*{\Scale}[2][4]{\scalebox{#1}{$#2$}}%

\title{Hierarchical Conditional Semi-paired Image-to-image Translation for Multi-task Image Defect Correction on Shopping Websites}
\name{Author(s) Name(s)\thanks{Thanks to XYZ agency for funding.}}
\address{Author Affiliation(s)}
\twoauthors
  {Moyan Li}
	{University of Michigan\\
	Ann Arbor, USA}
  {Jinmiao Fu, Shaoyuan Xu, Huidong Liu, Jia Liu, Bryan Wang}
	{Amazon\\
	Seattle, USA}

\begin{document}
%
{\maketitle}
\begin{abstract}
  On shopping websites, product images of low quality negatively affect customer experience.
  Although there are plenty of work in detecting images with different defects,
  few efforts have been dedicated to correct those defects at scale. A major challenge is that there are thousands of product types and each has specific defects, 
  therefore building defect specific models is unscalable. In this paper, we propose a unified Image-to-Image (I2I) translation model to correct multiple defects across different product types. Our model leverages an attention mechanism to hierarchically incorporate high-level defect groups and specific defect types to guide the network to focus on defect-related image regions. Evaluated on eight public datasets, our model reduces the Frechet Inception Distance (FID) by 24.6\% in average compared with MoNCE, the state-of-the-art I2I method. Unlike public data, another practical challenge on shopping websites is that some paired images are of low quality. Therefore we design our model to be semi-paired by combining the L1 loss of paired data with the cycle loss of unpaired data.
  Tested on a shopping website dataset to correct three image defects, our model reduces (FID) by 63.2\% in average compared with WS-I2I, the state-of-the art semi-paired I2I method.
\end{abstract}
\begin{keywords}
Image to Image translation, Computer Vision, Image defect auto-correction
\end{keywords}
\vspace{-2mm}

\section{Introduction}
\label{sec:intro}
\vspace{-2mm}
On shopping websites, product images provide customers with visual perception on the appearance of the products, thus play a critical role for customers’ shopping decisions. 
However, images provided by sellers usually contain various kinds of defects, such as non-white background or watermark. Given the definition of different defects and the corresponding training data (defective and non-defective images), 
it is straightforward to build an image classification model 
to detect defective images. However, 
re-shooting defect-free images is expensive and time-consuming for most sellers, which motivates us to build an ML framework to correct image defects automatically.

Image-to-image (I2I) translation is a promising approach because it can transform images from a source domain (defective) to a target domain (non-defective). 
I2I has a wide range of applications, such as image synthesis \cite{regmi2018cross,zhu2020sean}, 
semantic segmentation \cite{isola2017image, li2021semantic}, image inpainting \cite{yeh2016semantic, yu2018generative}, etc. I2I algorithms can be broadly classified into three categories: paired I2I \cite{park2019semantic, isola2017image}, unpaired I2I \cite{zhu2017unpaired, kim2017learning, lin2019learning, park2020contrastive, zhan2022modulated, hu2022qs}, and semi-paired I2I \cite{shukla2019extremely, tripathy2018learning}. In paired I2I, 
each image in the source domain is paired with an image in the target domain.
In unpaired I2I, data from both domains are available but not paired.
Semi-paired I2I uses both paired and unpaired data. 

There are two major limitations of I2I methods for the image defects correction on shopping websites:
First, I2I methods mainly focus on transforming images from one source domain to one or multiple target domains. They cannot support a single model that transforms multiple source domains to their corresponding target domains. On shopping websites, there are thousands of product types (e.g., shirt) and each has specific defects (e.g., non-white background).
Using the existing I2I methods, we need to train thousands of models to cover all the defects, which is unscalable. Second, defects on shopping websites are usually local, i.e., the defects only occupy a certain proportion of the whole image. 
 Most local I2I methods, such as InstaGAN \cite{mo2018instagan} or InstaFormer \cite{kim2022instaformer}, require mask or bounding box labels, which are expensive to obtain for all the defects.
 Although there are I2I methods \cite{kim2019u, tang2021attentiongan, zhang2019self} that use attention modules to identify local regions without requiring labels, they do not perform well when the images are from multiple product types with different defects. 
 

To enable the correction of multiple image defects with a single model while accurately locating the defect-related local regions, we design a hierarchical attention module. The module leverages the high-level defect groups (e.g., background related) and specific defect types (e.g., non-white background and watermark are two specific defects within the background-related group) to guide the generator to focus on the defect-related regions when trained on images of different defects and product types.
Furthermore, on shopping websites, due to the lack of human-audited paired data, 
the proposed model leverages synthesized paired data. To improve the model's robustness against synthesized pairs of low quality, we design our model to be semi-paired by adopting cycle loss on unpaired data.
In summary, our contributions include: (1) We propose a unified I2I pipeline that can correct multiple image defects across different product types, (2) We hierarchically inject the high-level defect groups and specific defect types
using attention modules,
which guides the network to focus on defect-related local regions,
(3) The proposed model can consume both paired data and unpaired data, while being robust against the synthesized pairs of low quality,
(4) The proposed model achieves better FID scores on eight public datasets and a shopping website dataset compared with the state-of-the-art I2I methods. 
  \vspace{-2mm}
\section{Method}
\begin{figure}[ht]
  \centering
  \includegraphics[width=0.95\linewidth]{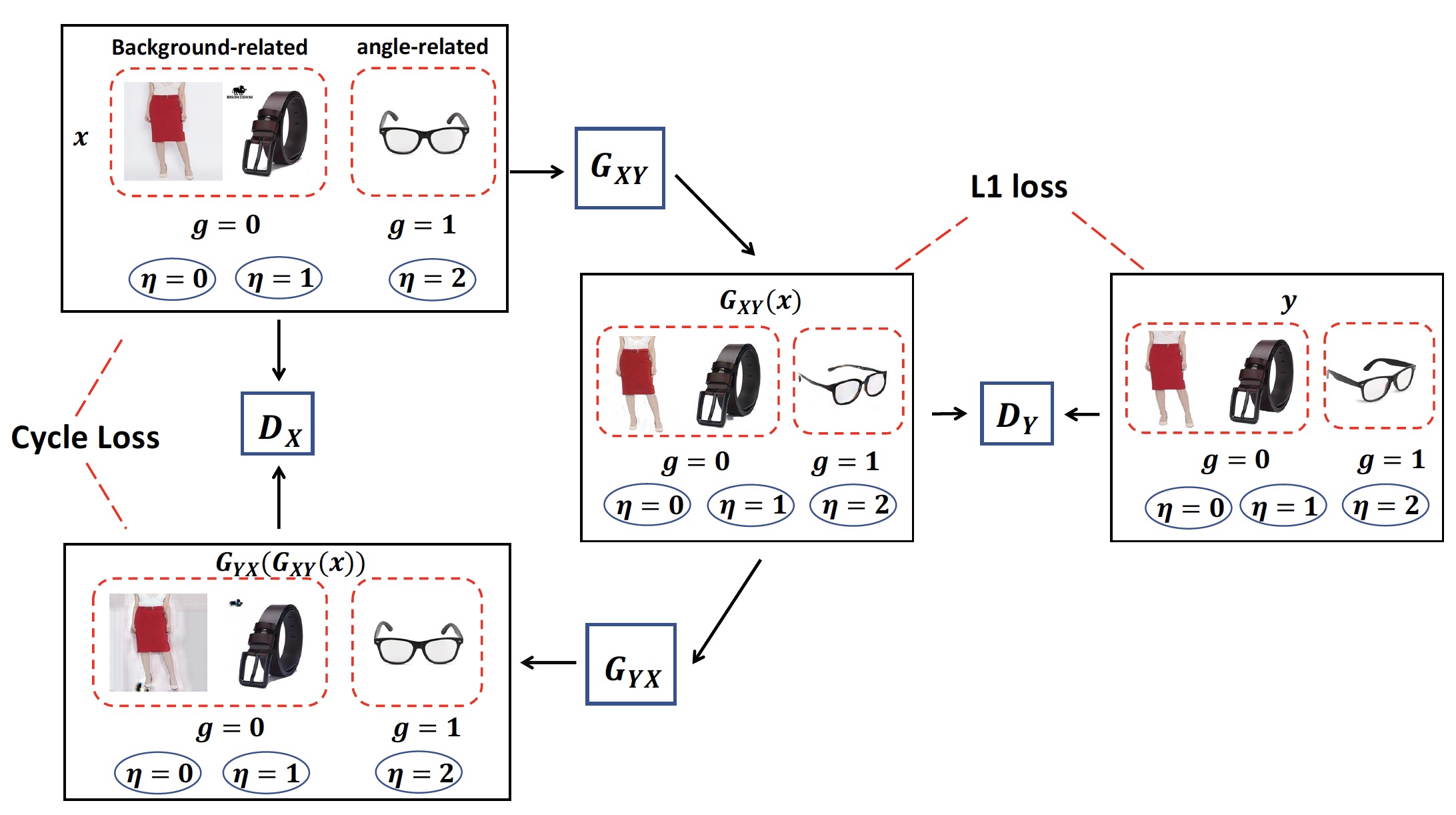}
  \caption{The architecture of the proposed framework. The discriminators $D_Y$ and $D_X$ leverage the L1 loss and cycle loss as the training objective for paired and unpaired images respectively. The generators $G_{XY}$ and $G_{YX}$ leverage the high level defect group $g$ and the specific defect type $\eta$ to locate defect related regions.}
  \label{fig: model}
  \vspace{-1mm}
\end{figure}

\vspace{-2mm}
Our proposed framework is shown in Figure \ref{fig: model}. It contains two generators $G_{XY}$ and $G_{YX}$ to transform images from domain $X$ to $Y$ and from Y to X respectively. For each pair of images $(x, y)$ from domain $X$ and $Y$, we train a discriminator $D_Y$ to distinguish $y$ and the transformed image $G_{XY}(x)$ using L1 loss. For each unpaired image $x$ in domain $X$, we train a discriminator $D_X$ to distinguish $x$ and $G_{YX}(G_{XY}(x))$, the generated image by transforming $x$ to domain $Y$ and further transforming it back to $X$. We insert the high level defect group $g$ and specific defect type $\eta$ to guide the two generators to focus on the defect related regions.
 \vspace{-1mm}
 
\vspace{-3mm}
\begin{figure}[ht]
  \centering
  \includegraphics[width=\linewidth]{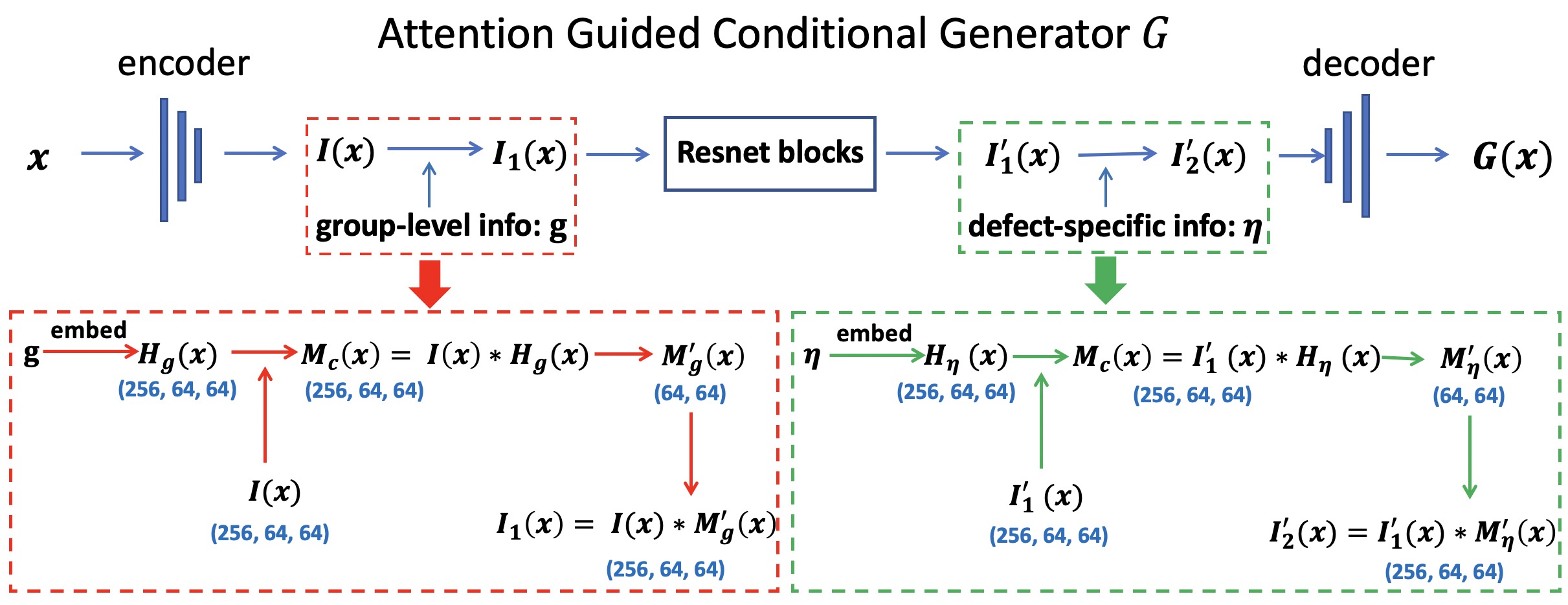}
  \caption{Attention-guided conditional generator structure. The encoder first maps an image $x$ into feature map $I(x)$. Then the high-level group label $g$ is embedded into the model as shown in the red box. After several Resnet blocks, we embed the specific defect type into the model as shown in the green box. Finally, the decoder will output the generated image.}
  \label{fig: generator}
  \vspace{-3mm}
\end{figure}
  \vspace{-3mm}
\subsection{Attention Guided Conditional Generator}
\vspace{-2mm}
To cover multiple defects with a single model, we propose an attention-guided conditional generator as shown in Fig.\ref{fig: generator}. It hierarchically consumes the high-level defect group and specific defect type. For example, 
image defects such as non-white background and watermark are background-related and incorrect sunglasses angle is angle-related. The high-level defect group $g$ (0 = background-related, 1 = angle-related) helps the generator decide whether to focus on distinguishing the background and foreground, or identifying the angle of the object. Furthermore, although non-white background and watermark belong to the same group, watermark usually only occupies a small region. Adding specific defect type $\eta$ (0 = non-white background, 1 = watermark, 2 = incorrect sunglasses angle) further helps the model tell the difference of the intra-group tasks. Inspired by \cite{ma2020fine}, the red block in Fig. \ref{fig: generator} illustrates the procedure of inserting $g$ into the model. Let $I(x)\in \mathbb{R}^{c^{\prime} \times h \times w}$ denote the feature map of an image $x$, 
we first project $g$ into a $c'$-dimensional vector through a linear transformation with $tanh$ activation, and perform spatial duplication to broadcast it into the same dimension as $I(x)$. 
Subsequently, we use $M_c(x) \in \mathbb{R}^{c' \times h \times w}$ computed as $M_c(x)=\tanh \left(\operatorname{Conv}_{1}(I(x) \odot H_g(x))\right)$ to measure the relevance of each element in $I(x)$ to $g$, 
where $\odot$ is the element-wise multiplication, and $\operatorname{Conv}_{1}$ is a convolution layer with 1 × 1 kernel. Then $M_g^\prime(x) \in \mathbb{R}^{h \times w}$, the summation of $M_c(x)$ across all the channels, represents the relevance of each spatial position in $I(x)$ to $g$. 
Finally, we denote $I_1(x) \in \mathbb{R}^{c' \times h \times w}$ as the element-wise multiplication between each channel of $I(x)$ and $M_g^\prime(x)$ as the updated feature map, which scales each position of $I(x)$ by its relevance to $g$. The similar process of inserting specific defect type $\eta$ is shown in the green box. 
The conditional labels $g$ and $\eta$ enable the model to cover multiple defects while focusing on the defect-related image regions.
  \vspace{-2mm}
\subsection{Semi-paired Structure}
\vspace{-2mm}
\label{subs: semi}
Another challenge on shopping websites is the lack of high-quality paired data. Although we can synthesize paired data, we cannot always guarantee the quality. For example, given an image of non-white background, we can synthesize its paired image of white background by detecting the foreground object and changing the background color to white. However, the synthesized image is not accurate if the foreground object is semi-transparent, of similar color of the background, or placed on another object (see the first line in Fig \ref{fig: amazon2}). The I2I model trained using such paired data will memorize the patterns of those low-quality pairs. To mitigate such effect, we leverage the cycle loss of unpaired data to ensure that the transformed images can be converted back.

  \vspace{-2mm}
\subsection{Training Loss}
\vspace{-2mm}
The training loss consists of the following components. First, adversarial losses in Eq.\ref{eq: adv_loss1} and Eq.\ref{eq: adv_loss2} ensure the generated images look realistic. Specifically, we adopt the relativistic discriminator \cite{jolicoeur2018relativistic}, i.e., $D_{Y}^{R e l}(y_1, y_2)=\operatorname{sigmoid}(C(y_1)-C(y_2))$, to measure the probability that $y_1$ looks more realistic than $y_2$, where $C$ refers to the non-transformed output of the discriminator. Eq.\ref{eq: adv_loss1} trains the discriminator to favor a real image $y$ against a generative image $\hat{y}$, while Eq.\ref{eq: adv_loss2} trains the generator to generate a $\hat{y}$ that looks more realistic than $y$.
\vspace{-2mm}
\begin{equation}\label{eq: adv_loss1}
    \Scale[0.83]{\begin{aligned}
        &\mathcal{L}_{G A N}^{D}\left(G_{X Y}, D_{Y}^{R e l}, X, Y\right)=\mathbb{E}_{y, \hat{y}}\left[\log D_{Y}^{R e l}(y, \hat{y})\right]
    \end{aligned}}
    \vspace{-2mm}
\end{equation}
\begin{equation}
\label{eq: adv_loss2}
    \Scale[0.83]{\begin{aligned}
        &\mathcal{L}_{G A N}^{G}\left(G_{X Y}, D_{Y}^{R e l}, X, Y\right)=\mathbb{E}_{y, \hat{y}}\left[\log D_{Y}^{R e l}(\hat{y}, y)\right]
    \end{aligned}}
    \vspace{-0.5mm}
\end{equation}
We denote the sum of these two losses as $\mathcal{L}_{GAN}^{XY}$, i.e., the total adversarial loss from domain $X$ to $Y$. Similarly, we denote $\mathcal{L}_{GAN}^{YX}$ as the total adversarial loss from $Y$ to $X$. 
Second, for paired data, we use reconstruction loss (Eq.\ref{eq: recon}) to minimize the distance between an image in the target domain and the image generated from the paired image in the source domain. 
\vspace{-1mm}
\begin{equation}\label{eq: recon}
    \begin{aligned}
        \Scale[0.83]{\mathcal{L}_{L 1}\left(G_{X Y}, G_{Y X}\right) =\mathbb{E}_{x} \|\left(G_{X Y}(x)-y \|_{1}\right.+\mathbb{E}_{y}\left\|G_{Y X}(y)-x\right\|_{1}}
    \end{aligned}
    \vspace{-2mm}
\end{equation}

For unpaired data, we use cycle loss (Eq.\ref{eq: cycle}) to ensure the generated images can be transformed back to the source domain, which prevents the model from overfitting to the paired images of low quality. 
\vspace{-2mm}
\begin{equation}\label{eq: cycle}
\Scale[0.8]{
    \begin{aligned}
        \mathcal{L}_{\mathrm{cyc}}(G_{XY}, G_{YX}) &=\mathbb{E}_{x}\left[\|G_{YX}(G_{XY}(x))-x\|_{1}\right] \\
        &+\mathbb{E}_{y}\left[\|G_{XY}(G_{YX}(y))-y\|_{1}\right]
    \end{aligned}}
    \vspace{-2mm}
\end{equation}
Besides, we add identity loss (Eq. \ref{eq: idt}), which encourages the generators to be close to an identity mapping when images from the target domain are fed to the generators \cite{rosca2017variational, zhu2017unpaired}. 
\vspace{-1mm}
\begin{equation}\label{eq: idt}
        \Scale[0.83]{\mathcal{L}_{i d t}\left(G_{X Y}, G_{Y X}\right) =\mathbb{E}_{x} \|\left(G_{Y X}(x)-x \|_{1}\right. +\mathbb{E}_{y}\left\|G_{X Y}(y)-y\right\|_{1}}
        \vspace{-1mm}
\end{equation}

At last, the total loss is defined as 
\vspace{-1mm}
 \begin{equation}
 \label{eq: loss}
     \Scale[0.83]{\mathcal{L}_{total}=\lambda_{1} \left(\mathcal{L}_{GAN}^{XY} +\mathcal{L}_{GAN}^{YX}\right) +\lambda_{2} \mathcal{L}_{L 1}+\lambda_{3} \mathcal{L}_{c y c l e}+\lambda_{4} \mathcal{L}_{i d t}}
 \end{equation}

  \vspace{-4mm}
\section{Experiments}
\subsection{Datasets}
\subsubsection{Public Datasets}
\label{subsec: public_data}
\vspace{-2mm}

We evaluate our model on eight public datasets, including four paired datasets: (\romannum{1}) maps: 1,296 map-to-aerial photo paired images \cite{isola2017image}, (\romannum{2}) facades (FA): 606 facade-to-segmentation paired images \cite{tylevcek2013spatial}, (\romannum{3}) edges2shoes (E2S): 50K paired images from UT Zappos50K dataset \cite{yu2014fine}, (\romannum{4}) edges2handbags (E2H): 20K Amazon Handbag images \cite{zhu2016generative}, and four unpaired datasets: (\romannum{1}) horse2zebra (H2Z): 1,267 horse images and 1,474 zebra images \cite{deng2009imagenet}, 
(\romannum{2}) apple2orange (A2O): 1,261 apple images and 1,529 orange images \cite{deng2009imagenet}, 
monet2photo (M2P): 1,193 Monet's paintings and 7,038 photos \cite{zhu2017unpaired}, (\romannum{4}) vangogh2photo (V2P): 800 Vangogh's paintings and 7,038 photos \cite{zhu2017unpaired}.
We split each dataset into training and test (80/20). We combine the four paired datasets where we assign maps and FA to the segmentation-related group ($g$ = 0), and E2S and E2H to the colorization-related group ($g$ = 1). Similarly, we combine the four unpaired datasets where we assign H2Z and A2O to the color-related group ($g$ = 0), and M2P and V2P to the style-related group ($g$ = 1).
\vspace{-2mm}

\subsubsection{Image Defects Dataset}
\label{subsec: amazon_data}
\vspace{-2mm}
We collected images with three different defects including non-white background (non-Wbg), watermark (WM) and incorrect sunglasses angle (in-SA) from a shopping website. For non-Wbg and WM images, we use salient object detection \cite{lee2022tracer} to detect the objects and clean up the background to construct paired images of white background or no watermarks. We construct the pairs of in-SA images using the main and secondary image of each sunglasses product. In total there are 2,703 non-Wbg image pairs, 4,465 WM image pairs and 8,070 in-SA pairs. We assign these three specific defect types two high level defect groups where non-white background and watermark belongs to the background-related group ($g=0$) and incorrect sunglasses angle belongs to the angle-related group ($g=1$). 

\vspace{-2mm}
\subsection{Training Details}
\vspace{-2mm}

We resize each image to 256 x 256. For both datasets, we train the models for 300 epochs using Adam optimizer [12] with batch size 1. Following Pix2Pix \cite{isola2017image}, we set an initial learning rate of 0.0002, which is fixed for first 150 epochs and decays linearly to zero afterwards. We set $\lambda_1$, $\lambda_2$, $\lambda_3$, $\lambda_4$ to be 1, 150, 10 and 10 following WS-I2I \cite{shukla2019extremely}. WS-I2I is a semi-paired I2I model which leverages the same loss functions as in this paper. The difference is WS-I2I cannot cover multiple tasks due to the lack of guidance from $g$ and $\eta$. We use Frechet Inception Distance (FID) \cite{heusel2017gans} as the evaluation metric, which measures the distance between the distributions of generated images and real images. Lower FID score means better performance. 

\vspace{-2mm}
\subsection{Performance Comparison}
\label{subs: performance}
\begin{figure}[htbp]
  \centering
  \includegraphics[width=0.9\linewidth]{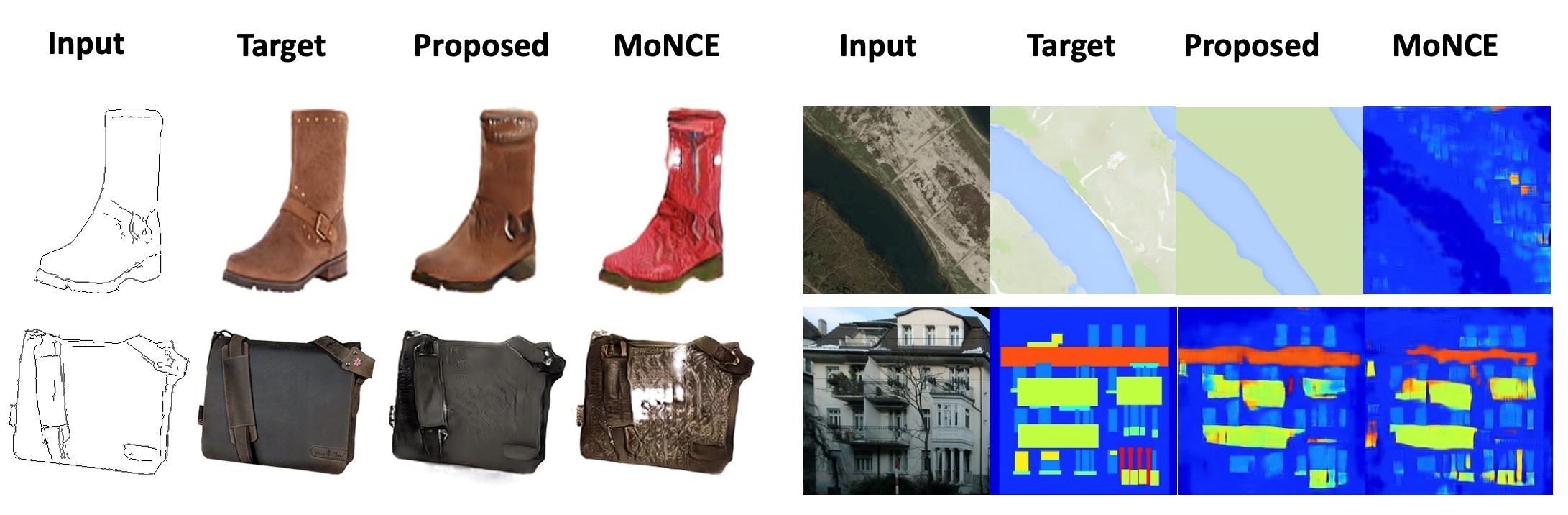}
  \caption{The comparison between the proposed model and MoNCE on the combined four paired public data.}
  \label{fig: paired}
  \vspace{-5mm}
\end{figure}

\begin{figure}[ht]
  \vspace{-5mm}
  \centering
  \includegraphics[width=0.9\linewidth]{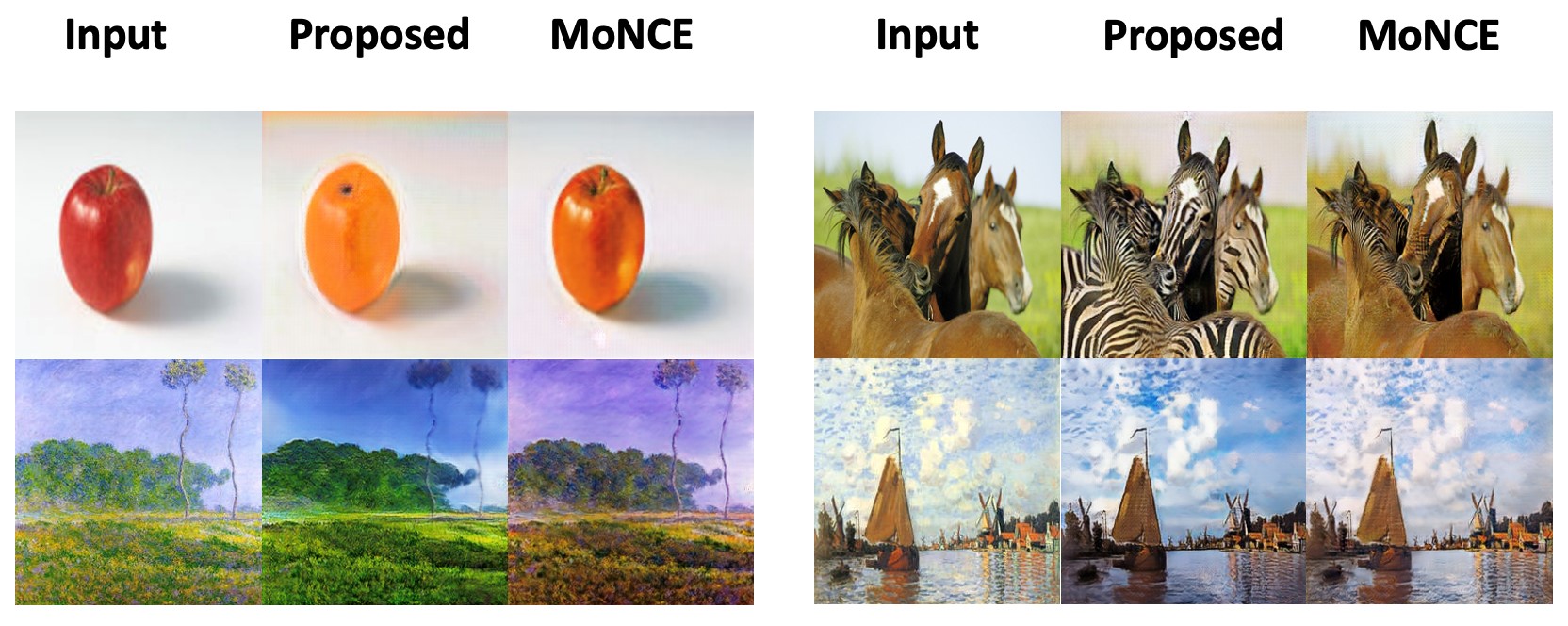}
  \caption{The comparison between the proposed model and MoNCE on the combined four unpaired public data.}
  \label{fig: unpaired}
    \vspace{-5mm}
\end{figure}

\vspace{-2mm}
On the combined four paired public data, we compare the proposed model with the baseline method  Pix2Pix\cite{isola2017image} and the state-of-the-art method MoNCE \cite{zhan2022modulated}. Since there are no unpaired data, we set $\lambda_3$ and $\lambda_4$ to $0$. Table \ref{tab:public} (Paired) shows that our method is consistently better than Pix2Pix and MoNCE. As shown Figure \ref{fig: paired}, when trained on multiple tasks, MoNCE tends to transform an image from the source domain of a task to the target domain of another task (e.g., transforming a google map to a segmentation instead of an aerial-photo) due to the lack of guidance from $g$ and $\eta$.
On the combined four unpaired public data, we compare our method with the baseline method CycleGAN \cite{zhu2017unpaired} and MoNCE \cite{zhan2022modulated}. Since there are no paired data, we set $\lambda_2$ to $0$. We observe similar pattern as the paired data as shown in Table \ref{tab:public} (Unpaired) and Fig \ref{fig: unpaired}. 

\begin{table}
  \footnotesize
  \caption{FID scores of the paired (maps, FA, E2S and E2H) and unpaired (H2Z, A2O, M2P and V2P) I2I translation  using different types of models. The baseline model use Pix2Pix for paired tasks and CycleGAN for unpaired tasks.}
  \label{tab:public}
  \begin{tabular}
{p{10mm}|p{4mm}|p{3.5mm}|p{4.5mm}|p{4.5mm}|p{4.5mm}|p{4.5mm}|p{4.5mm}|p{4.5mm}}
    \toprule
    \multicolumn{1}{c|}{}&
    \multicolumn{4}{c|}{Paired} &
    \multicolumn{4}{|c}{Unpaired} \\
    \cmidrule{2-9}
    \quad &Map &FA & E2S& E2H &H2Z &A2O &M2P& V2P\\
    \midrule
    Baseline & 301 & 266 & 166 & 111& 185 & 288 & 197 & 141\\
    MoNCE & 200 & 198 & 131 & 99 & 135 & 203 & 178 & 135\\
    \textbf{Proposed} & \textbf{107} & \textbf{155} & \textbf{86} & \textbf{89} & \textbf{90} & \textbf{170} & \textbf{145} & \textbf{113}\\
  \bottomrule
\end{tabular}
\vspace{-2mm}
\end{table}

\begin{figure}[ht]
  \centering
  \vspace{-5mm}
  \includegraphics[width=0.7\linewidth]{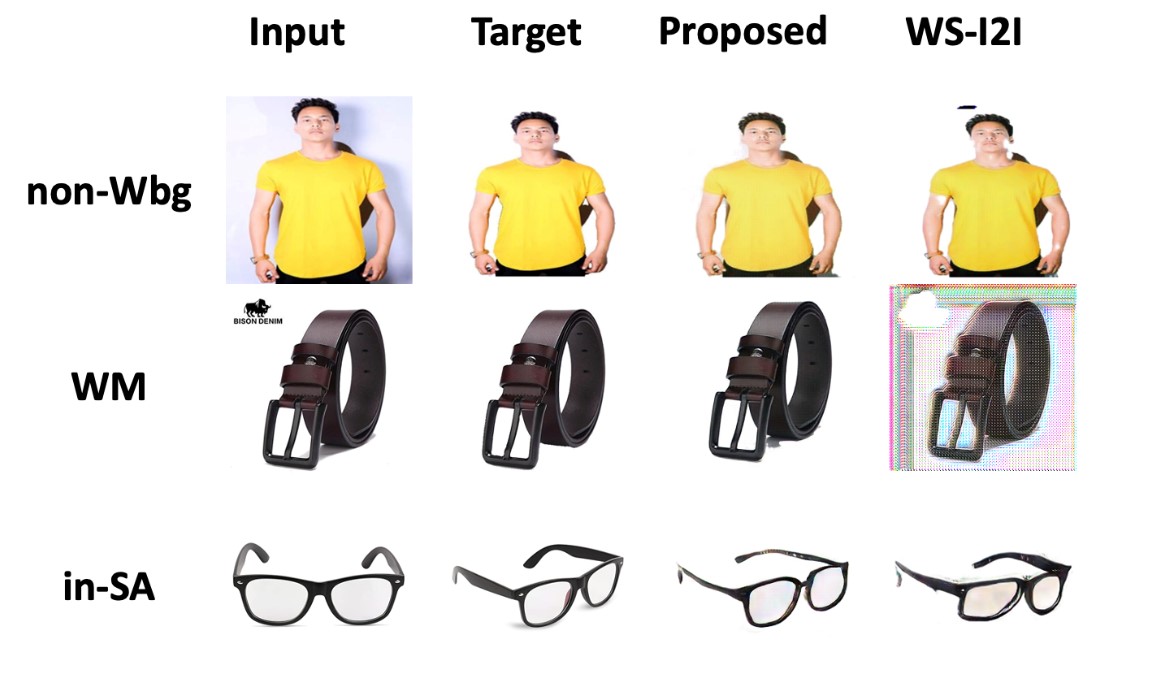}
  \caption{Comparison between the proposed model with WS-I2I on the Image Defects Detection dataset.}
  \label{fig: amazon1}
  \vspace{-5mm}
\end{figure}

On the Image Defects Dataset to correct three image defects including non-Wbg, WM and in-SA, to avoid the model from memorizing the patterns of the low-quality synthesized pairs of non-Wbg and WM images, we configure our model to be semi-paired by combining the reconstruction loss (Eq. \ref{eq: recon}) of paired data with the cycle loss (Eq. \ref{eq: cycle}) and the identity loss (Eq. \ref{eq: idt}) of non-Wbg and WM images. Therefore instead of MoNCE \cite{zhan2022modulated}, we compare the proposed model with WS-I2I \cite{shukla2019extremely}, the state-of-the-art semi-paired I2I model to the best of our knowledge. As shown in Table \ref{tab: amazon_result}, our proposed model performs better in all three tasks. Fig \ref{fig: amazon1} shows some examples where our proposed model can transform the images correctly while WS-I2I cannot. 

\begin{table}
\footnotesize
  \caption{FID scores of the proposed model and WS-I2I on correcting non-Wbg, WM and in-SA on the Image Defects Dataset}
  \label{tab: amazon_result}
  \begin{tabular}{p{15mm}|p{12mm}|p{6mm}|p{8mm}}
    \toprule
    \quad & non-Wbg &WM & in-SA\\
    \midrule
   \textbf{proposed} & \textbf{22} & \textbf{20} & \textbf{66} \\
     WS-I2I & 55& 89& 138\\
    \bottomrule
  \end{tabular}
  \vspace{-2mm}
\end{table}

\subsection{Ablation Study}
\begin{figure}[ht]
\vspace{-5mm}
  \centering
  \includegraphics[width=\linewidth]{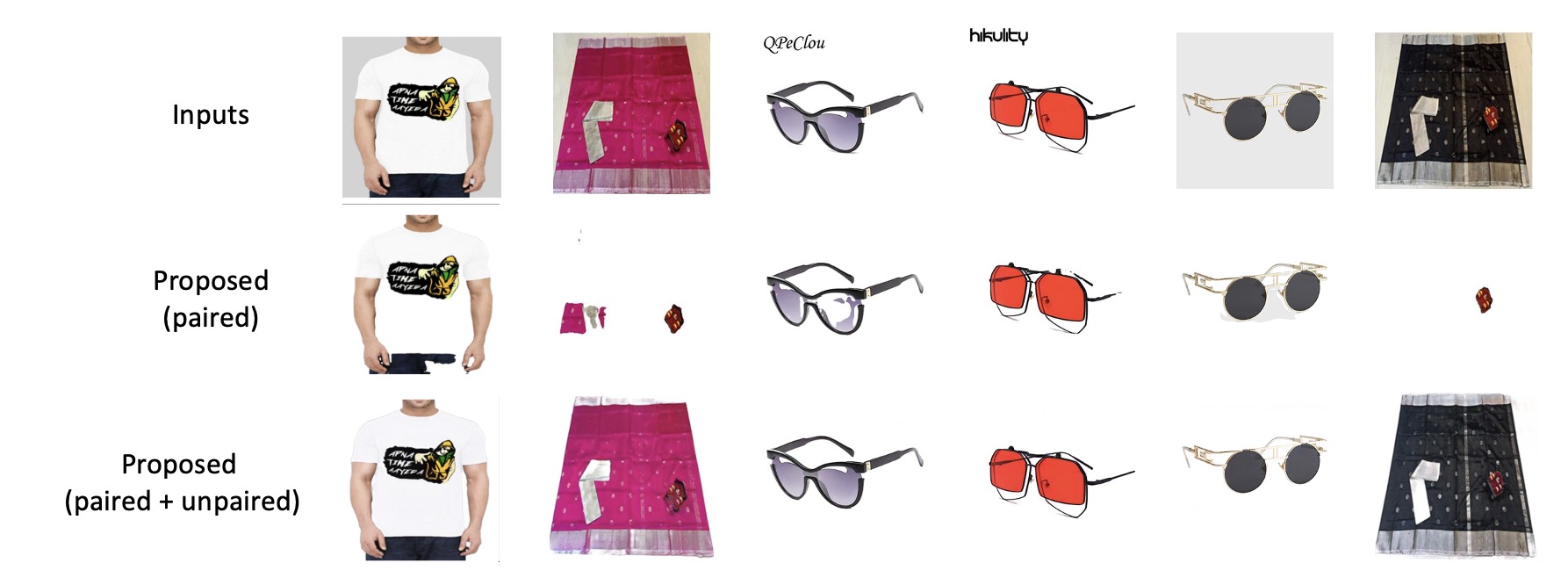}
  \caption{Visualization of results on the Image Defects Dataset to compare the proposed model and the proposed model w/o unpaired data.}
  \label{fig: amazon2}
  \vspace{-5mm}
\end{figure}

\vspace{-2mm}
We conduct the ablation study in Table \ref{tab: ablation1} to demonstrate the effect of adding high-level group $g$, specific defect type $\eta$ and using unpaired data. 
We can conclude that hierarchically adding $g$ and $\eta$ into the model can significantly improve the model performance (by comparing the FID scores in the first line with the second and third line in Table \ref{tab: ablation1}). The results also demonstrate the benefit of leveraging unpaired data (by comparing the first line and the fourth line in Table \ref{tab: ablation1}).
We visualize the benefit of using unpaired data when some of the paired data are of low quality in Fig \ref{fig: amazon2}. When the object is semi-transparent, of similar color of the background, or placed on another object (first line in Fig \ref{fig: amazon2}), the synthesized paired images will be of low quality, which will negatively affect the model performance if trained using only paired data (second line in Fig \ref{fig: amazon2}). In comparison, our proposed model (third line in Fig \ref{fig: amazon2}) is robust against such situation.

\section{Conclusion}
\vspace{-2mm}
In this paper, we propose a unified Image-to-Image translation framework to transform images from multiple source domains to their corresponding target domains. By hierarchically injecting high-level defect groups and specific defect types using attention modules, the framework can capture different levels of image defect patterns for better defect correction. The model is semi-paired so that it is robust against the low-quality paired data. Our framework is scalable to multiple image defects in various domains and can significantly improve FID compared to the state-of-the-art I2I methods. 


\vspace{-5mm}

\begin{table}
\vspace{-5mm}
  \footnotesize
  \caption{FID scores of the proposed model, the proposed model w/o $g$, the proposed model w/o $g$ and $\eta$, and the proposed model w/o unpaired data on correcting non-Wbg, WM and in-SA in the Image Defects Dataset}
  \label{tab: ablation1}
  \begin{tabular}{p{26mm}|p{15mm}|p{7mm}|p{10mm}}
    \toprule
    \quad & non-Wbg &WM & in-SA\\
    \midrule
   \textbf{proposed} & \textbf{22} & \textbf{20} & \textbf{66} \\
     w/o $g$ & 27 & 25 & 70 \\
     w/o $g$, $\eta$ & 55& 89& 138\\
     w/o unpaired data & 43& 53& 68\\
    \bottomrule
  \end{tabular}
  \vspace{-4mm}
\end{table}

\clearpage
\bibliographystyle{IEEEbib}
\bibliography{refs}

\end{document}